\documentclass{article}
\pdfoutput=1
\usepackage{amsmath}
\usepackage{times}
\usepackage{graphicx}
\usepackage{color}
\usepackage{multirow}
\usepackage{caption}
\usepackage{subcaption}
\usepackage{placeins}
\usepackage{url}

\usepackage{rotating}
\usepackage{latexsym}

\DeclareGraphicsExtensions{.png,.pdf}
\graphicspath{{imglocal/}{img/}{./}} 
\title{Stochastic Interpretation of Quasi-periodic Event-based Systems}
\author{Hesham Mostafa and Giacomo Indiveri\\
  Institute for Neuroinformatics\\
  University of Zurich and ETH Zurich\\
  \{hesham,giacomo\}@ini.uzh.ch
}

\begin{document}
\date{}
\maketitle

\thispagestyle{empty}
\markboth{}{NC instructions}




%


%
%
\begin{abstract}
Many networks used in machine learning and as models of biological neural networks make use of stochastic neurons or neuron-like units. We show that stochastic artificial neurons can be realized on silicon chips by exploiting the quasi-periodic behavior of mismatched analog oscillators to approximate the neuron's stochastic activation function. We represent neurons by finite state machines (FSMs) that communicate using digital events and whose transitions are event-triggered. The event generation times of each neuron are controlled by an analog oscillator internal to that neuron/FSM and the frequencies of the oscillators in different FSMs are incommensurable. We show that within this quasi-periodic system, the transition graph of a FSM can be interpreted as the transition graph of a Markov chain and we show that by using different FSMs, we can obtain approximations of different stochastic activation functions. We investigate the quality of the stochastic interpretation of such a deterministic system and we use the system to realize and sample from a restricted Boltzmann machine. We implemented the quasi-periodic event-based system on a custom silicon chip and we show that the chip behavior can be used to closely approximate a stochastic sampling task. 

\end{abstract}

\section{Introduction}
Stochastic spiking neurons are often used in biological network models to account for the variability of neural responses~\cite{holt_etal96}. Stochastic neural responses can have important computational implications such as allowing neural networks to sample from probability distributions~\cite{Buesing_etal11} or to transmit small subthreshold input signals~\cite{longtin93}. In machine learning, networks using neuron-like units with stochastic activation functions are often used to realize probabilistic generative models of input data~\cite{Smolensky86,salakhutdinov_hinton09}. Many multi-layer networks used in classification and encoding tasks~\cite{Hinton_Salakhutdinov06,vincent_etal10,Hinton_etal06} also make use of stochastic neuron elements to limit the amount of information that flows from one layer to another during unsupervised pre-training. 

An open question is how such stochastic networks can be realized efficiently on custom silicon chips. Custom chip implementations of either machine learning network architectures~\cite{Pham_etal12} or the more biologically inspired spiking networks~\cite{Merolla_etal14b} can offer significant performance gains compared to simulating these networks on conventional general-purpose CPUs or GPUs. Straightforward implementations of stochastic networks  would use an explicit true- or pseudo-random noise source in each unit to realize uncorrelated fluctuations. 

We propose here an efficient, distributed, and easily implementable scheme for the generation of largely uncorrelated fluctuations in a large number of neuron elements. The scheme exploits the non-repeating phase relations in a quasi-periodic system. Each neuron element has access to the state of an analog oscillator. Due to the inevitable inhomogeneities in the silicon fabrication process, the different oscillators are guaranteed to be incommensurable, i.e, have oscillation frequencies that are not rational multiples of each other. The phase relations between these oscillators varies irregularly in an aperiodic manner. By developing the communication scheme between the neuron elements so that the interaction strength between a group of neurons depends both on the weights between them as well as the phase relations within the group, we can obtain neural activity that changes in a non-repeating manner and that can be modelled in stochastic terms.



In this paper we present a stochastic formulation that matches the 'statistics' of this quasi-periodic system and that replaces each deterministic neuron with an equivalent neuron having a stochastic activation function. We investigate the fidelity of this approximation and evaluate how the quasi-periodic system performance differs from a system that uses high quality pseudo-random number generators. We focus on digital neurons where each neuron is a simple finite state machine (FSM). The neurons/FSMs communicate in an event based fashion.



We use this system to implement a prototypical stochastic network which is the restricted Boltzmann machine. We present measurements from a physical implementation of such a system on a custom VLSI chip to demonstrate that the mismatch inherent in a standard VLSI process is sufficient to make different instances of the the same oscillator circuit oscillate at different frequencies. We use the custom VLSI chip to implement a simple sampling task. 

\section{Sigmoidal units}
\begin{figure}[t]
\label{fig:sigmoid}
  \centering
  \begin{subfigure}[b]{0.4\textwidth}
    \includegraphics[width=\textwidth]{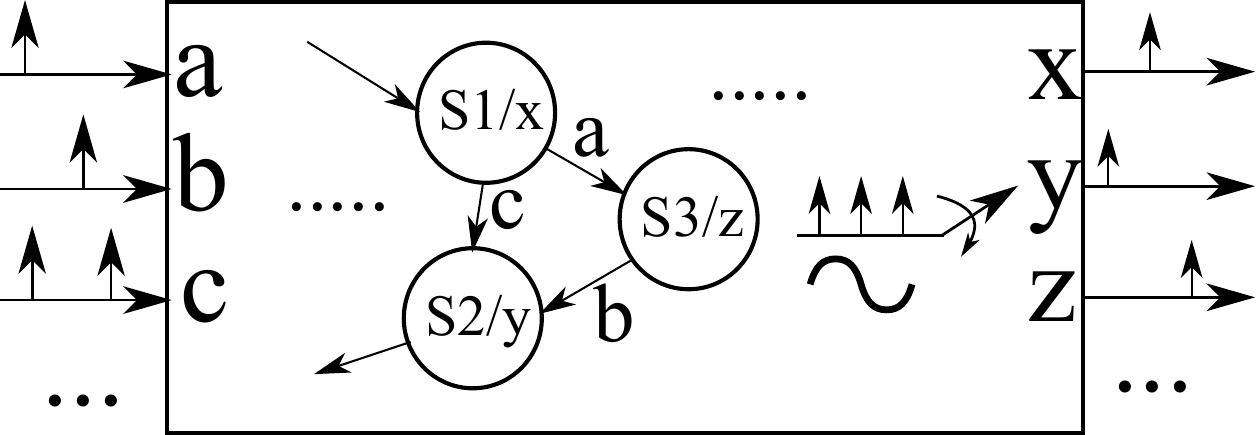}
    \subcaption{}
    \label{fig:sigmoid_a}
  \end{subfigure}
  \quad
  \begin{subfigure}[b]{0.45\textwidth}
    \includegraphics[width=\textwidth]{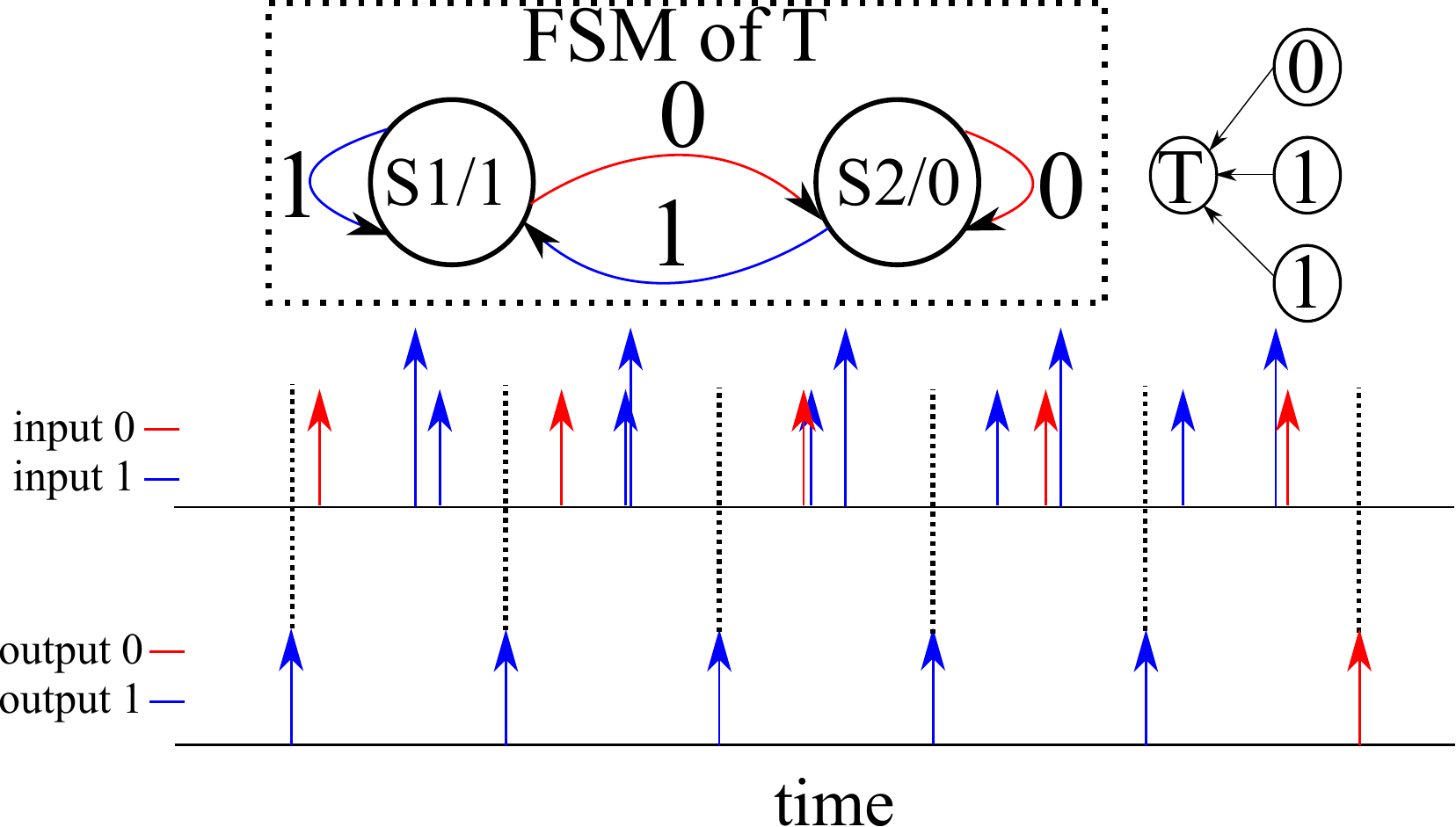}
    \subcaption{}
    \label{fig:sigmoid_b}
  \end{subfigure}
  \\
  \begin{subfigure}[b]{0.25\textwidth}
    \includegraphics[width=\textwidth]{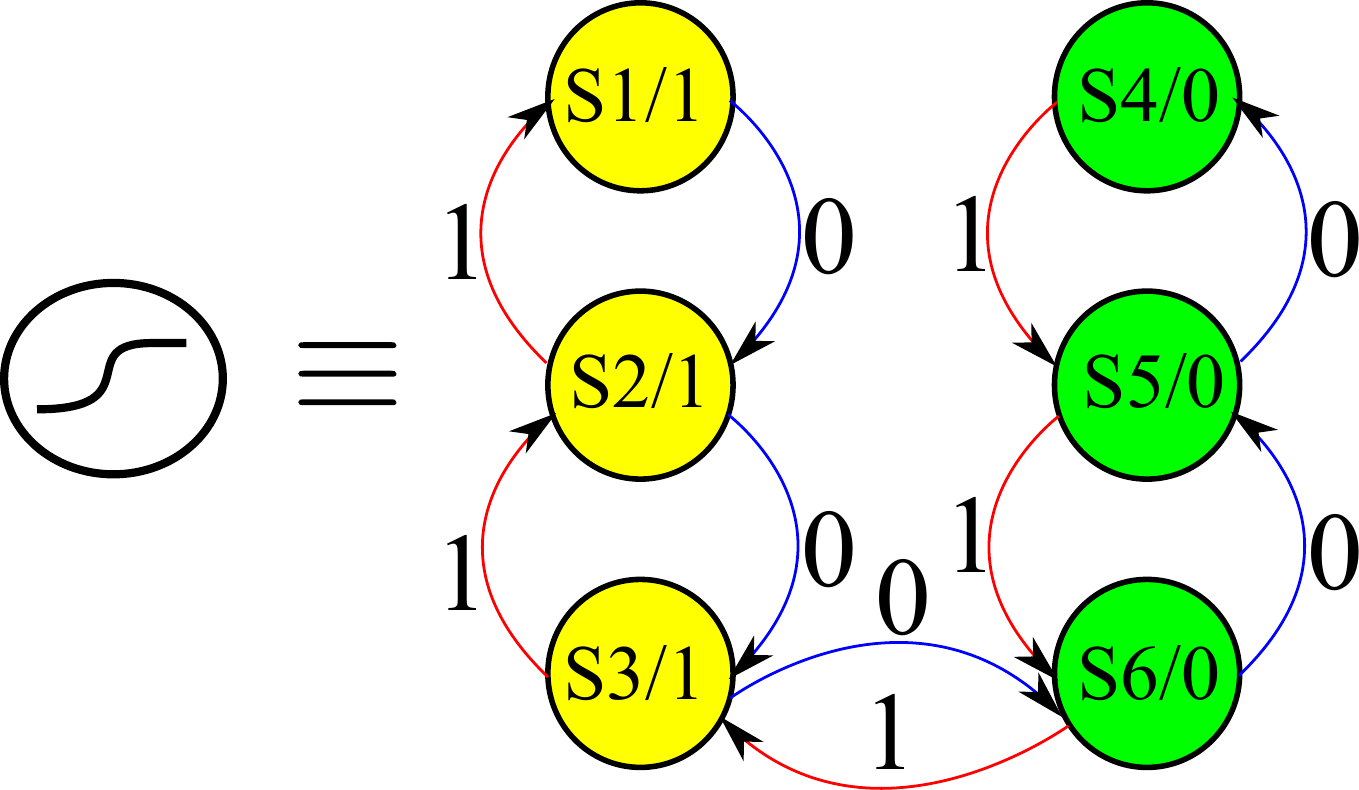}
    \subcaption{}
    \label{fig:sigmoid_c}
  \end{subfigure}
  \quad
  \begin{subfigure}[b]{0.25\textwidth}
    \includegraphics[width=\textwidth]{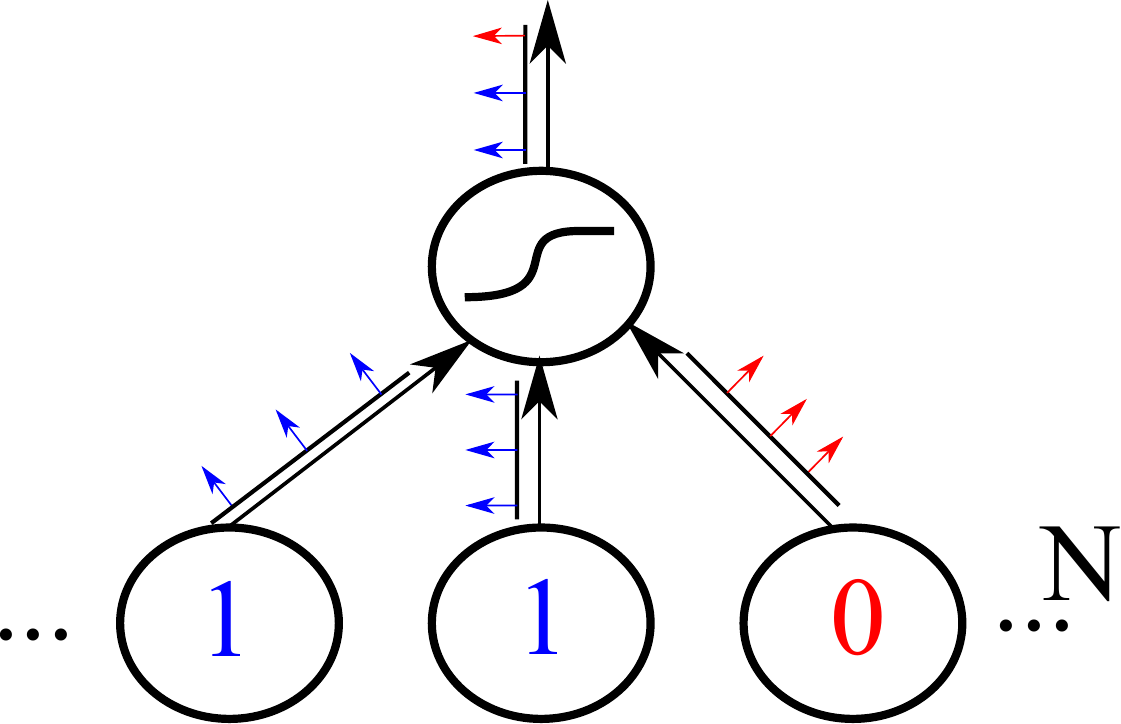}
    \subcaption{}
    \label{fig:sigmoid_d}
  \end{subfigure}
  \begin{subfigure}[b]{0.45\textwidth}
    \includegraphics[width=\textwidth]{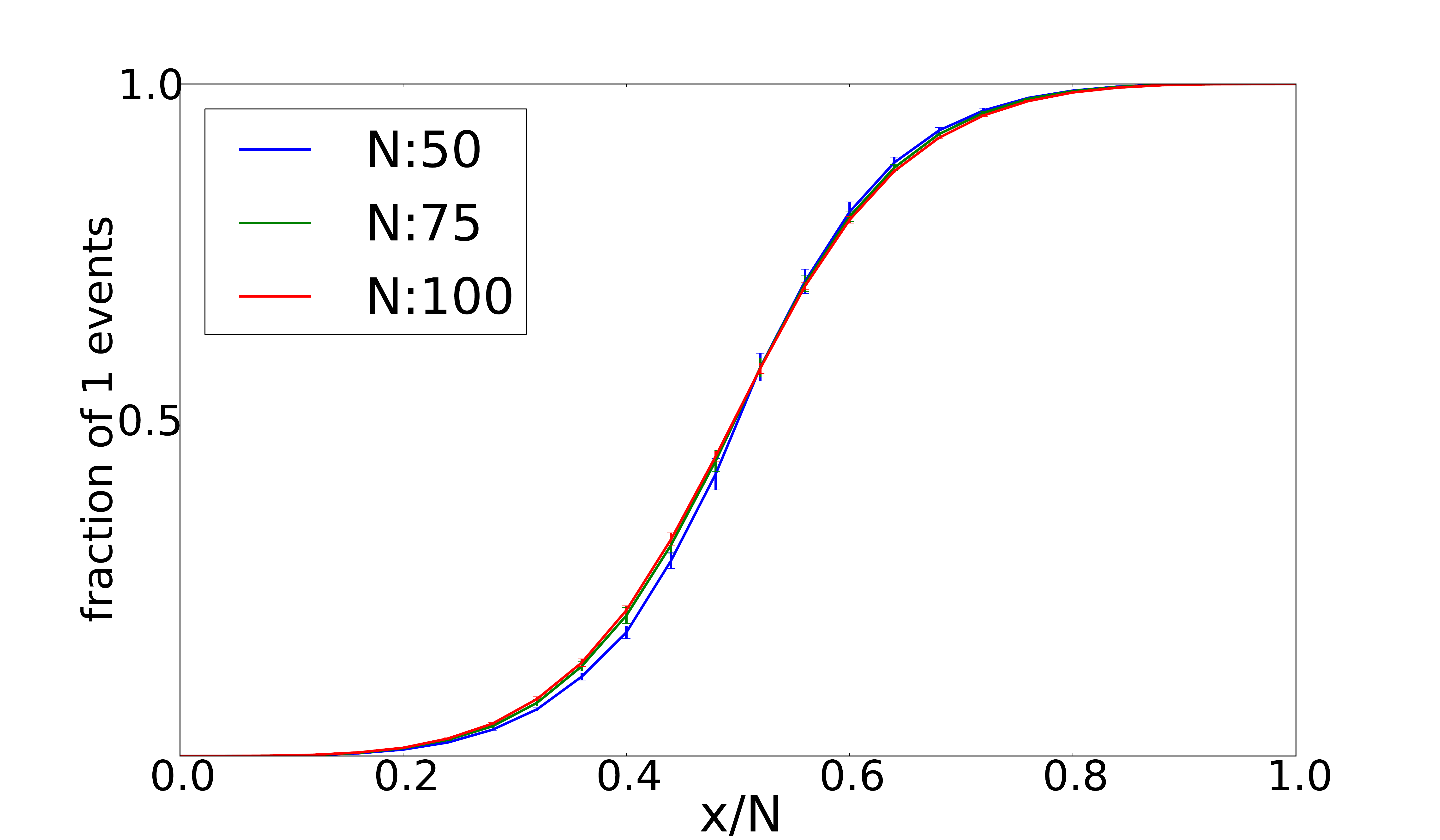}
    \subcaption{}
    \label{fig:sigmoid_e}
  \end{subfigure}
  \\
  \begin{subfigure}[b]{0.45\textwidth}
    \includegraphics[width=\textwidth]{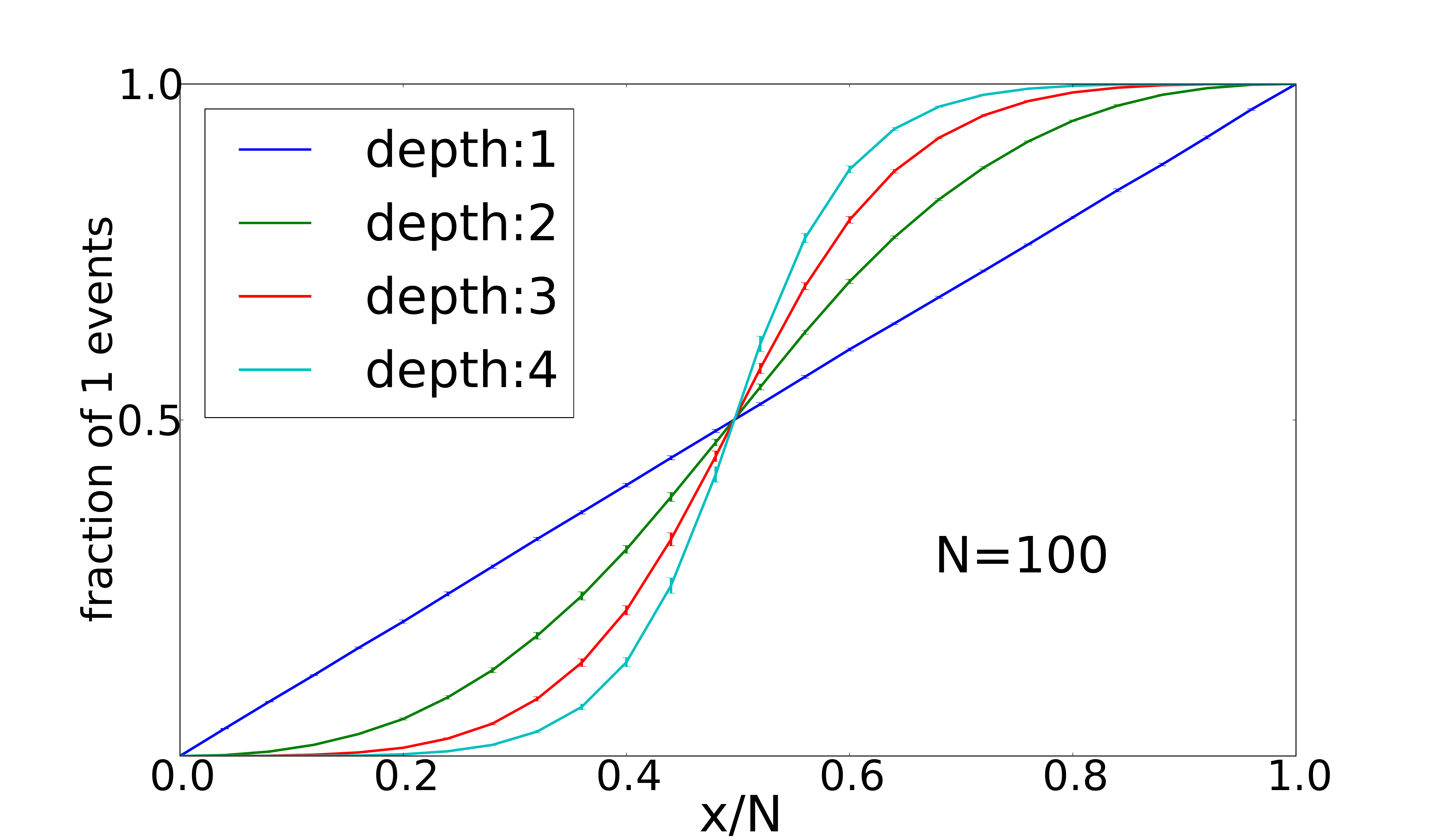}
    \subcaption{}
    \label{fig:sigmoid_f}
  \end{subfigure}
  \quad
  \begin{subfigure}[b]{0.45\textwidth}
    \includegraphics[width=\textwidth]{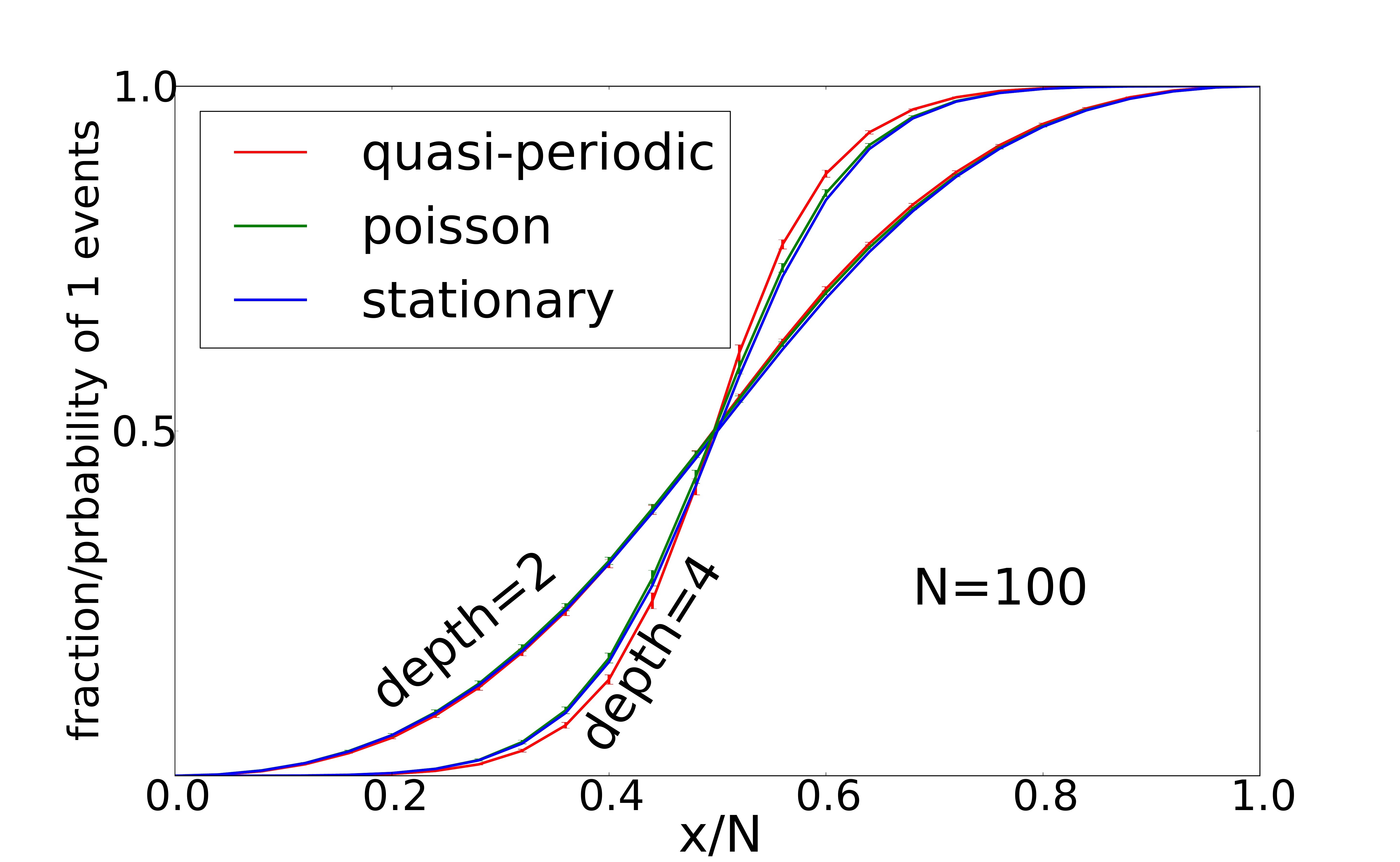}
    \subcaption{}
    \label{fig:sigmoid_g}
  \end{subfigure}

  \caption{ (\subref{fig:sigmoid_a}) General form of a neuron element. Arrival of input events trigger FSM transitions. When an event is generated by the internal oscillator, it is routed to one output port selected based on the current state of the neuron. (\subref{fig:sigmoid_b}) A simple two state neuron element and an illustrative simulation. (\subref{fig:sigmoid_c}) A neuron with a counter-like FSM. (\subref{fig:sigmoid_d}) The neuron from  \subref{fig:sigmoid_c} receiving events from $x$ neurons with value $1$ and $N-x$ neurons with value $0$. (\subref{fig:sigmoid_e}) Fraction of $1$ events generated by the target neuron in \subref{fig:sigmoid_d} after $1.5*10^5$ events as a function of $x/N$ (the fraction of afferent $1$ neurons). (\subref{fig:sigmoid_f}) Shape of the target neuron activation function crucially depends on the depth of its counter/FSM. (\subref{fig:sigmoid_g})Activation functions when each neuron generates a periodic train of events (quasi-periodic) or a Poisson train (Poisson), or calculated directly from the stationary distribution of the neuron-equivalent Markov chain (stationary). Lines in \subref{fig:sigmoid_e}, \subref{fig:sigmoid_f}, and \subref{fig:sigmoid_g} are averages over 5 trials. Error bars are the standard deviation. Oscillator frequencies redrawn in each trial uniformly from the range $[40,50]$\,Hz.    }
\end{figure}

Figure~\ref{fig:sigmoid_a} shows the general structure of a neuron. It is composed of a FSM in which events arriving on the inputs ports $a,b,c,\ldots$ trigger state transitions. Associated with the FSM is an analog oscillator that generates a regular train of events. The neuron routes each event from this internal oscillator to one of the output ports. If the FSM is in state $Si/j$ when the analog oscillator generates an event, this event is routed to output port $j$. A neuron with $k$ output ports can transmit $log_2k$ bits of information in each oscillator cycle. We consider the index of the output port on which a neuron last generated an event as the neuron's current value. Neurons can be connected together (output ports to input ports) and we assume the oscillator frequencies in the different neurons are incommensurable. 

Consider the FSM of a simple neuron, neuron $T$, that is shown in the top part of Fig.~\ref{fig:sigmoid_b}. This neuron receives a stream of $1$ events from two neurons and a stream of $0$ events from one neuron. From the FSM of $T$, it is clear that $T$ will route the internal oscillator events to the $1$ ($0$) port if the last event  it received was $1$ ($0$) as shown in Fig.~\ref{fig:sigmoid_b}. Even though the values of the three source neurons are constant, eventually, the $0$ input event arrives just before the end of the cycle of $T$, and the value of $T$ changes from $1$ to $0$. The value of a neuron (the identity of its last output event) thus does not solely depend on the values of the source neurons but it also crucially depends on the order of arrival of their events. This order is continuously changing in an aperiodic manner due the incommensurable oscillation frequencies of the oscillators inside the neurons. 

Figure~\ref{fig:sigmoid_c} shows the FSM of a slightly more complex neuron that also has two input ports and two output ports. We refer to neurons having such a counter-like FSM as sigmoidal neurons. Assume this neuron is receiving events from $N$ constant-value source neurons as shown in Fig.~\ref{fig:sigmoid_d} where $x$ of the source neurons are sending $1$ events and $N-x$ are sending $0$ events. Figure~\ref{fig:sigmoid_e} shows that the fraction of $1$ events generated by the sigmoidal neuron of Fig.~\ref{fig:sigmoid_c} is a sigmoidal function of $x/N$ or the fraction of $1$ events impinging on the neuron (the neuron has to generate either a $1$ or a $0$ event for each event from its internal oscillator). The shape of the activation function in Fig.~\ref{fig:sigmoid_e} is robust to changes in $N$. It, however, crucially depends on the depth of the counter in the sigmoidal neuron as shown in Fig.~\ref{fig:sigmoid_f}. The FSM of the sigmoidal neuron shown in Fig.~\ref{fig:sigmoid_c} has depth $3$ while the FSM of the neuron in Fig.~\ref{fig:sigmoid_b} has depth $1$.

To accurately calculate the fraction of $1$ events generated by the target sigmoidal neuron, we need to enumerate all possible orderings of the $N$ source neurons' events within one cycle of the target neuron and calculate what fraction of these orderings will put the target neuron's FSM in one of the $1$ output states (the yellow states). The calculation will depend on the target neuron's initial state and is further complicated by the difference in oscillator frequencies which might lead to some source neurons not generating any events, or generating multiple events, during one cycle of the target neuron. The analysis can be greatly simplified if we interpret the system behavior in stochastic terms. At the heart of this stochastic interpretation is the following approximation:

{\centerline{\it  The periodic train of events generated by a neuron is treated as a Poisson train}}

This approximation is similar to the stochastic approximation previously used to analyze the behavior of quasi-periodic winner-take-all networks~\cite{Mostafa_etal15,Muller15}.
Under the Poisson event generation assumption, the probability at any time that the next input event is $1$ is $x/N$ (the fraction of $1$ source neurons), and the probability that the next input event is $0$ is $1-x/N$. The FSM of a neuron can thus be cast as a Markov chain having the same structure but with each $1$ edge replaced by an edge with transition probability $x/N$ and each $0$ edge by an edge with transition probability $1-x/N$. The probability of a particular output is the sum of the probabilities of the states yielding that output in the Markov chain stationary distribution. For each $x/N$ value, we can thus calculate the stationary distribution of the Markov chain and obtain the probability of generating a $1$ event. This is plotted in Fig.~\ref{fig:sigmoid_g} together with the activation functions of the deterministic quasi-periodic system and the stochastic system where the oscillators generate a Poisson (instead of a periodic) train of events. The stochastic approximation is valid but becomes less accurate as the depth of the sigmoidal neuron FSM/counter increases. That is because the state of a deeper counter reflects a longer history of events and in that longer history, differences between the periodic and the Poissonian event generation mechanisms become apparent. For instance, the Poissonian neuron can generates two events in quick succession which is impossible in the periodic case. 

\begin{figure}[t]
\label{fig:corr}
  \centering
  \begin{subfigure}[b]{0.45\textwidth}
    \includegraphics[width=\textwidth]{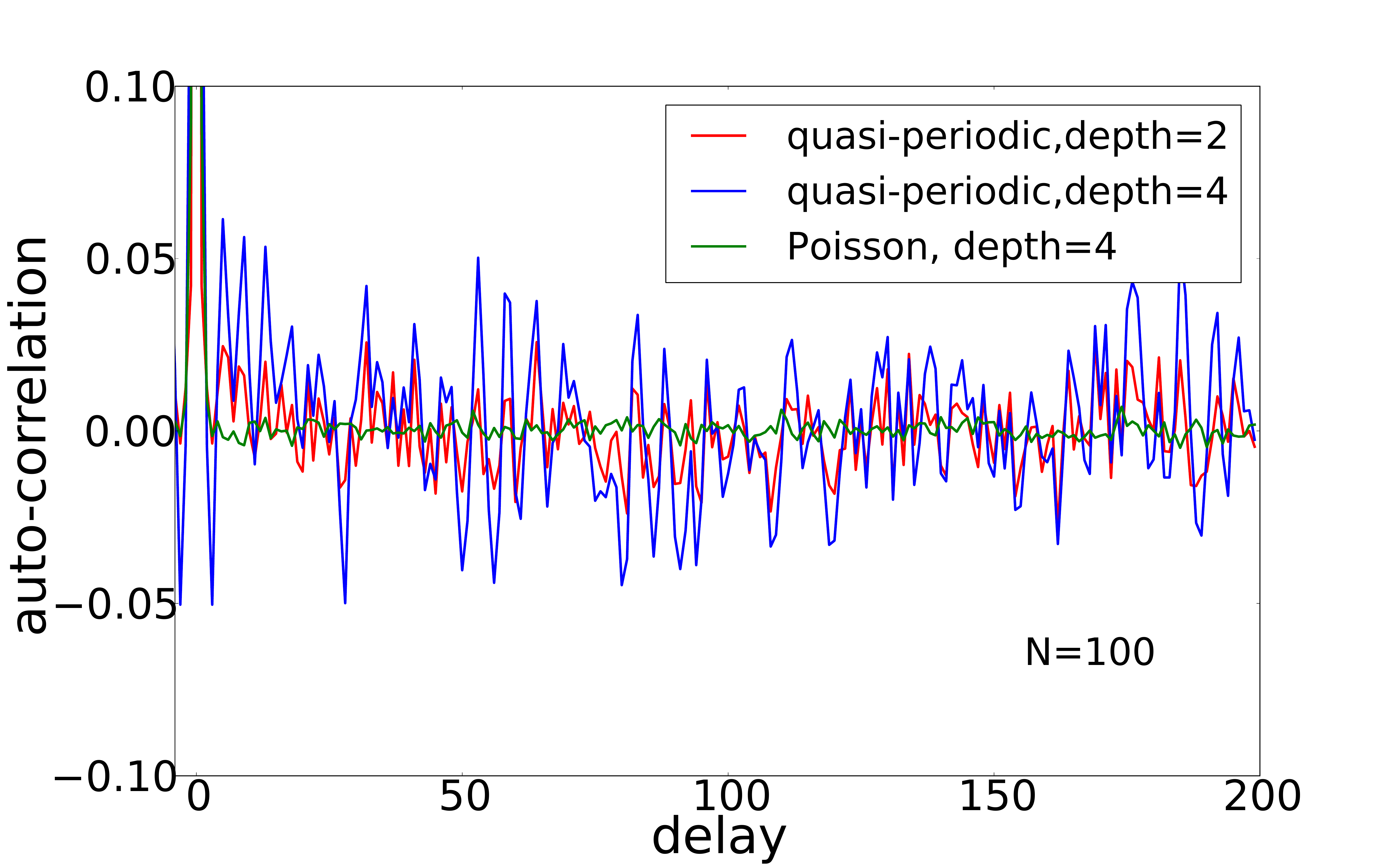}
    \subcaption{}
    \label{fig:corr_a}
  \end{subfigure}
  \quad
  \begin{subfigure}[b]{0.45\textwidth}
    \includegraphics[width=\textwidth]{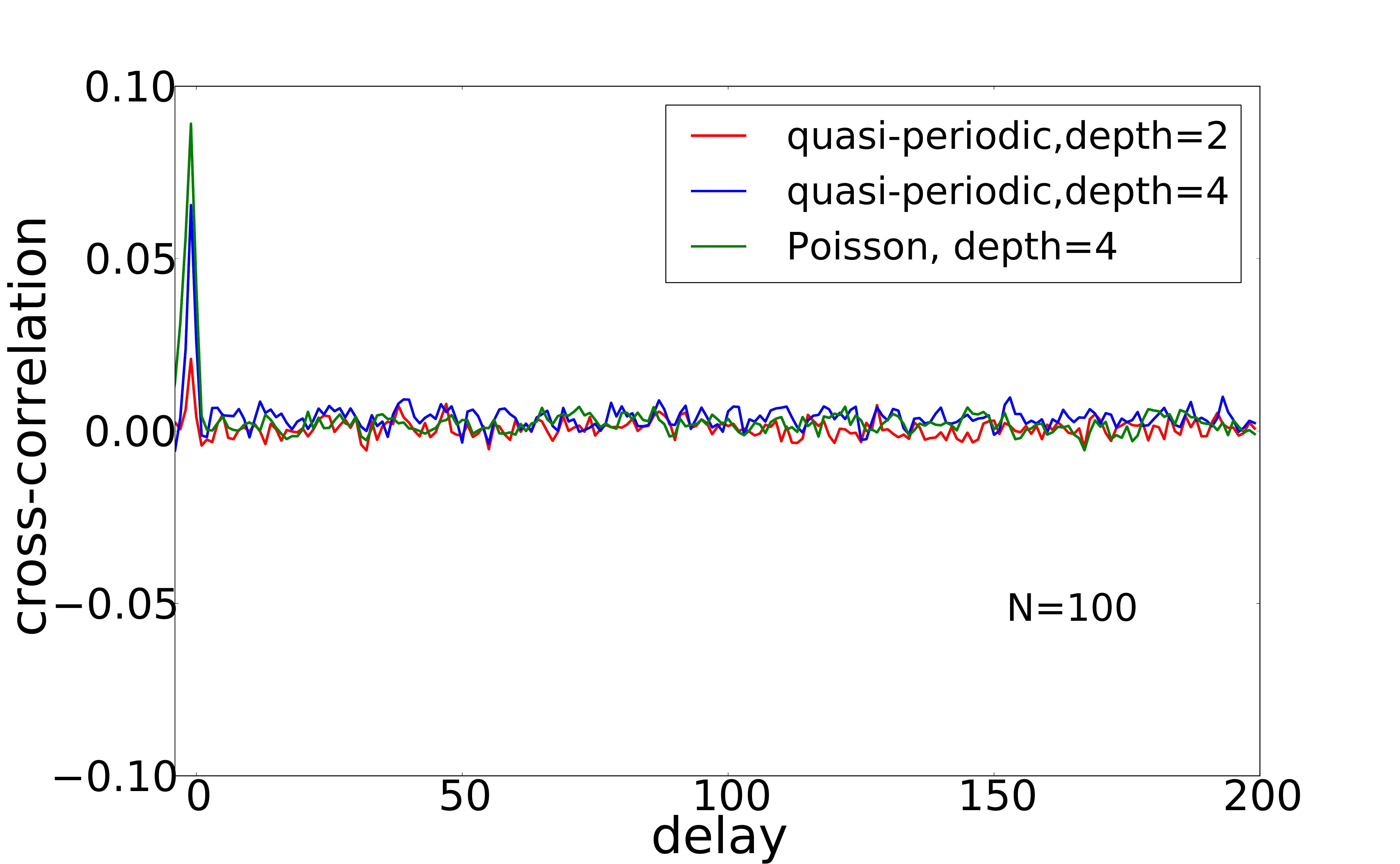}
    \subcaption{}
    \label{fig:corr_b}
  \end{subfigure}
\caption{(\subref{fig:corr_a}) Part of the auto-correlation of the sequence of $1.5*10^5$ bits generated by a target neuron when $x/N = 0.5$ in the periodic and Poissonian event generation cases . (\subref{fig:corr_b}) Cross-correlation of 2 sequences of bits generated by two target neurons receiving the same input when $x/N = 0.5$. Cross-correlation was calculated after adjusting the sequence of bits generated by one neuron so that the temporally closest events/bits in the two sequences occur in the same sequence position.}
\end{figure}

The sequence of bits/events generated by the target neuron in the quasi-periodic system in Fig.~\ref{fig:sigmoid_d} has small but non-decaying correlations as shown in Fig.~\ref{fig:corr_a}. A deeper counter (longer memory) in the target neuron results in higher correlations as it becomes more difficult for incoming events to yield a state/value at the cycle's end that is independent of the state at the cycle's beginning (which reflects the previous bit value). The non-decaying and non-repeating correlation structure is a product of the quasi-periodic nature of the system which causes the phase relations among the oscillators to almost repeat after a while and yield similar event orderings. Two target neurons having different frequencies and receiving events from the same neurons generate sequences of bits with small cross-correlation which is on par with the Poissonian system as shown in Fig.~\ref{fig:corr_b}. 

In summary, the quasi-periodic event-based system shown in Fig.~\ref{fig:sigmoid_d} admits a stochastic interpretation in which event generation in each neuron is assumed to be Poissonian instead of periodic. This interpretation works because of the finite memory in each neuron which renders its output sensitive to the order of arrival of input events. Since this order changes in an irregular manner, the target neuron can see it as ``random''. Empirically, we observe that the stochastic approximation is accurate for other types of neurons/FSMs as long as the number of source neurons is large compared to the number of states in the target neuron's FSM.

\subsection{Restricted Boltzmann Machines}
The behavior of a network of neurons, where each neuron has the form shown in Fig.~\ref{fig:sigmoid_c} can be interpreted in probabilistic sampling terms. The network connectivity (the way events are routed) induces a 'probability distribution' over the possible states of the network. In a Gibbs sampling fashion, we interpret each event from a neuron as a sample drawn from the distribution over this neuron's values conditioned on the current state of all other neurons. Under this interpretation, we show that the quasi-periodic network can reproduce the sampling behavior of a stochastic RBM. 

An RBM is a Markov random field on a bipartite graph of binary 0/1 units. The graph has $N_v$ visible units and $N_h$ hidden units. Visible and hidden units are bidirectionally  connected according to the $N_v*N_h$ weight matrix ${\it W}$. There are no connections between visible units or between hidden units. ${\bf x}$ and ${\bf y}$ are the visible and hidden bias vectors respectively. ${\bf v}$ and ${\bf h}$ are the vectors representing the states of the visible and hidden units respectively. The probability of a particular configuration is:
\begin{subequations}
\label{eq:rbm}
\begin{equation}
  P({\bf v},{\bf h}) = \frac{e^{\frac{-E({\bf v},{\bf h})}{T}}}{Z}  
\end{equation}
\begin{equation}
  E({\bf v},{\bf h}) = - {\bf x}^T{\bf v} - {\bf y}^T{\bf h}  - {\bf v}^T{\it W}{\bf h} \quad\quad  Z = \sum\limits_{{\bf v},{\bf h}}e^{\frac{-E({\bf v},{\bf h})}{T}}
\end {equation}
\end{subequations}
where $E({\bf v},{\bf h})$ is the energy of configuration $[{\bf v},{\bf h}]$, $T$ the model temperature, and $Z$ the normalizing constant or partition function. Generating samples from the distribution in Eq.~\ref{eq:rbm} is typically done through Gibbs sampling which updates a visible(hidden) unit conditioned on the current state of the hidden(visible) units by drawing a sample from:

\begin{equation}
\label{eq:sample}
P(v_i = 1 | {\bf h}) = \sigma_T(\sum\limits_{j=1}^{N_h}w_{ij}h_j + x_i ) \quad\quad P(h_j = 1 | {\bf v}) = \sigma_T(\sum\limits_{i=1}^{N_v}w_{ij}v_i + y_j)
\end{equation}
where $\sigma_T(x) = 1/(1+e^{-x/T})$ is the logistic sigmoid function. A hidden or visible unit/neuron thus needs to have a logistic sigmoid stochastic activation function so that its output corresponds to a Gibbs sampling update step. 

\begin{figure}[t]
\label{fig:rbm}
  \centering
  \begin{subfigure}[b]{0.45\textwidth}
    \includegraphics[width=\textwidth]{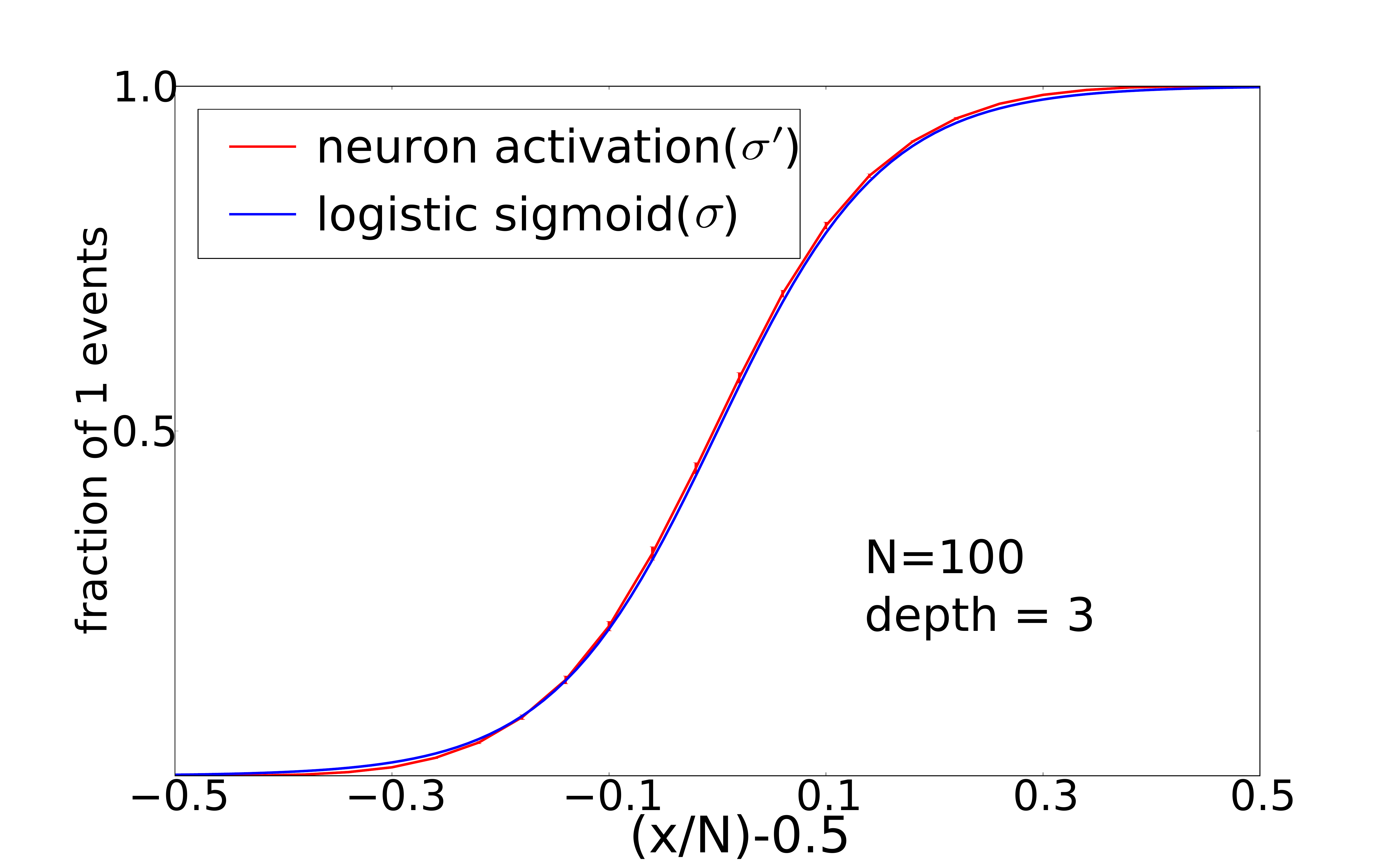}
    \subcaption{}
    \label{fig:rbm_a}
  \end{subfigure}
  \quad
  \begin{subfigure}[b]{0.27\textwidth}
    \includegraphics[width=\textwidth]{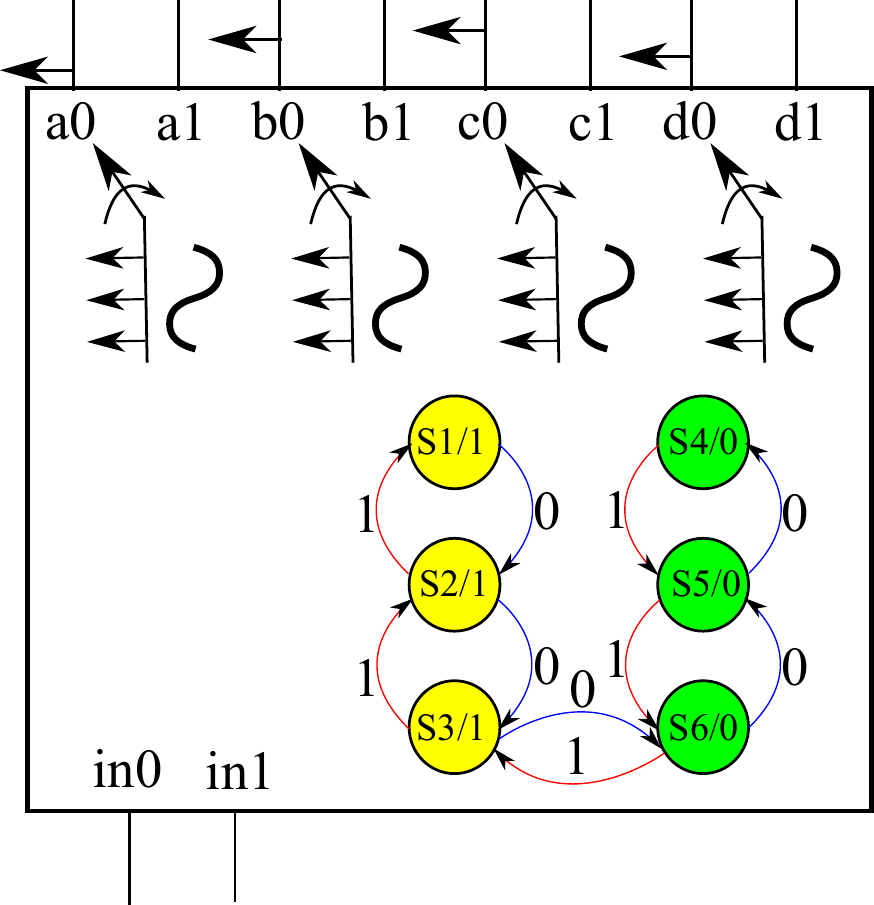}
    \subcaption{}
    \label{fig:rbm_b}
  \end{subfigure}
  \\
  \begin{subfigure}[b]{0.5\textwidth}
    \includegraphics[width=\textwidth]{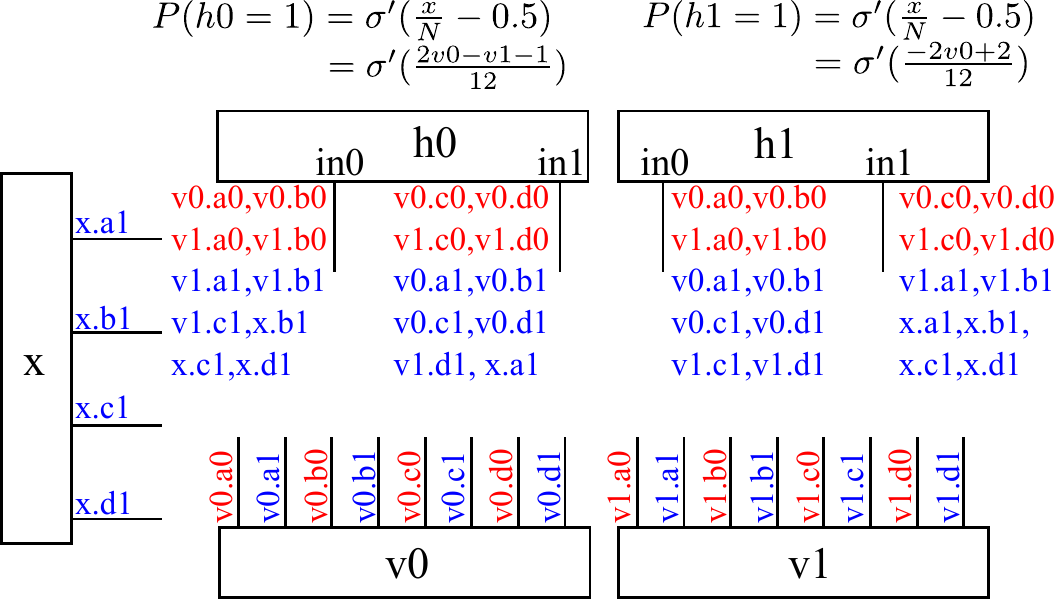}
    \subcaption{}
    \label{fig:rbm_c}
  \end{subfigure}
  \quad
  \begin{subfigure}[b]{0.45\textwidth}
    \includegraphics[width=\textwidth]{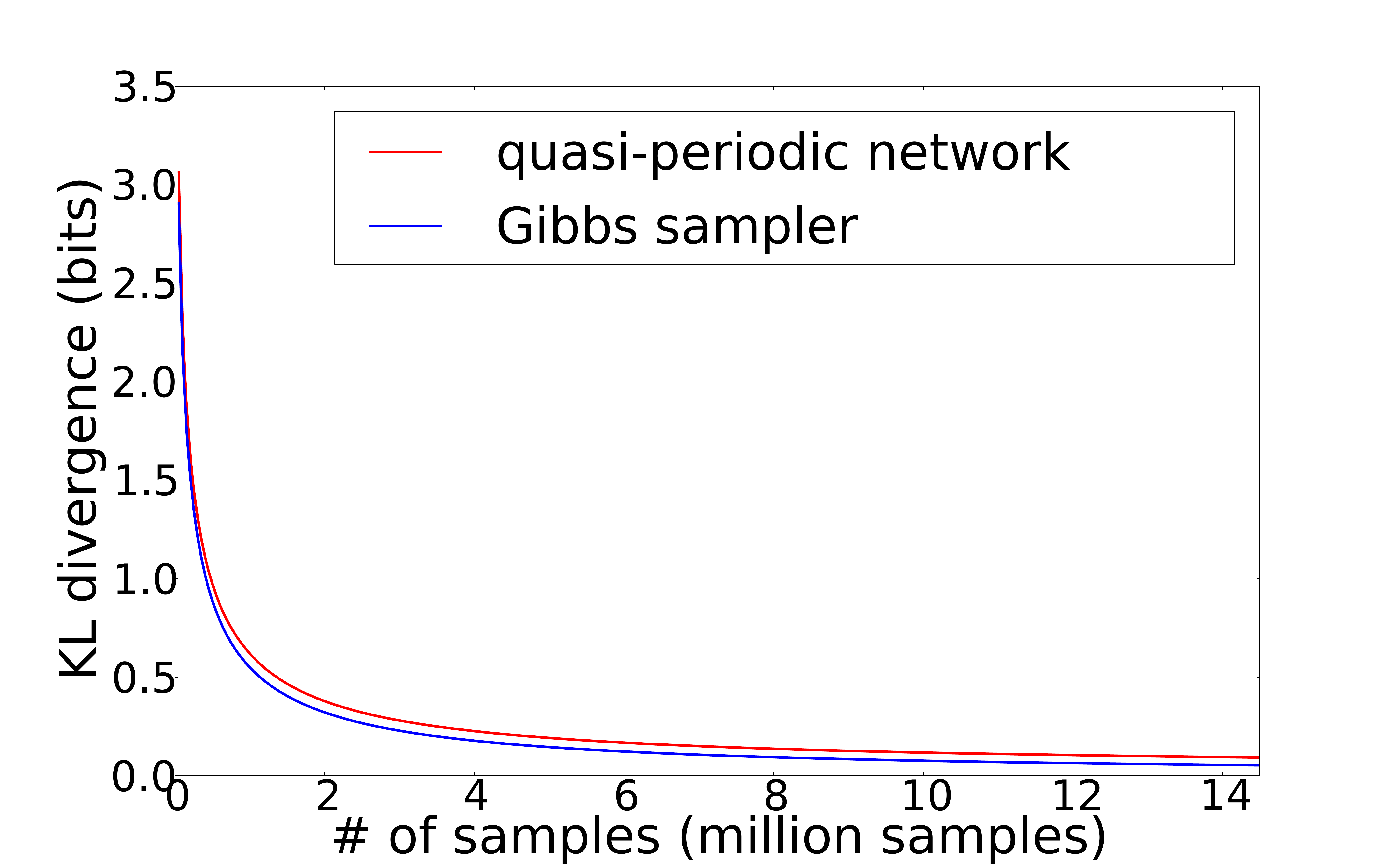}
    \subcaption{}
    \label{fig:rbm_d}
  \end{subfigure}

\caption{(\subref{fig:rbm_a}) The depth $3$ neuron activation as a function of the excess fraction of input $1$ events closely matches a logistic sigmoid function. (\subref{fig:rbm_b}) Implementation of a sigmoidal neuron with $4$ output event streams. (\subref{fig:rbm_c}) Example RBM with weights $w_{00}=2$, $w_{01}=-2$, $w_{10}=-1$, and $w_{11}=0$, and hidden biases $x_0 = -1$ and $x_1=2$. Only visible to hidden connections and hidden biases are shown. Next to each input port of a hidden unit is a list of the output ports of the visible units whose events are routed to that input port. $0$ output ports are in red, and $1$ output ports are in blue. The probability of generating a $1$ event for each hidden neuron is shown. $\sigma'$ is the sigmoidal neuron activation function shown in \subref{fig:rbm_a}. (\subref{fig:rbm_d}) $KL$ divergence between the sampling distribution and the true probability distribution of an RBM when the samples are generated by a Gibbs sampler (blue) and by the quasi-periodic network implementation of the RBM (red). Simulation was repeated four times with different random RBMs and yielded virtually identical curves.}
\end{figure}

Figure~\ref{fig:rbm_a} shows that the activation function of a sigmoidal neuron with depth $3$ closely matches that of a logistic sigmoid. As shown in Fig.~\ref{fig:sigmoid_e}, the shape of the activation function is robust to the number of source neurons. The sigmoidal neuron shown in Fig.~\ref{fig:sigmoid_c} could thus be used to represent a unit in an RBM. The sigmoidal neuron's activation function is plotted as a function of the excess fraction of incoming $1$ events: $x/N - 0.5$. For each neuron, we thus need to make the quantity $x/N - 0.5$ equal to a weighted sum of the states of the source units, plus a bias term as in Eq.~\ref{eq:sample}. 

Due to the discrete nature of the neurons, we can only use discrete weights and biases. To implement an RBM using $5$ possible weights/biases: $\{-2,-1,0,1,2\}$, we need to use the neuron shown in Fig.~\ref{fig:rbm_b} to implement each RBM unit. The neuron is similar to the one in Fig.~\ref{fig:sigmoid_c} except that it has $4$ independent oscillators that each generates a periodic train of events. The events from each oscillator can go to one of two output ports based on the state of the neuron's FSM. Thus, the neuron generates four event streams. If the neuron's state is $S1$, $S2$, or $S3$. The events from the $4$ oscillators are routed to output ports $a1$, $b1$, $c1$, and $d1$, otherwise they are routed to $a0$, $b0$, $c0$, and $d0$.  Figure ~\ref{fig:rbm_c} is an example of how to connect neurons/units to implement weighted connections with discrete weights. Only the visible to hidden connections (which are a mirror of the hidden to visible connections) and the hidden biases are shown. The four $0$ event streams from a visible unit are distributed equally on the $0$ and $1$ input ports of the target neurons. The four $1$ event streams of a visible unit are distributed on each target neuron's input ports so as to implement weighted connections. The connection scheme ensures that regardless of the state of the visible units, the number of incoming event streams at a hidden units ($N$) is always $12$. The hidden biases are implemented as weighted connections from an always $1$ neuron. 

It is easy to verify in Fig.~\ref{fig:rbm_c} that the excess fraction of incoming $1$ events is $(2v0-v1-1)/12$ and $(-2v0+2)/12$ for hidden units/neurons $h0$ and $h1$ respectively (the weights are scaled by a constant factor). Arbitrary RBMs with weights/biases in the range $\{-2,-1,0,1,2\}$ can be similarly implemented. 
We implemented an RBM with $10$ visible and $10$ hidden units whose weights and biases are drawn randomly from the integers between $-3$ and $3$ (Each neuron/unit has $6$ oscillators and $6$ output event streams) . The RBM is small enough to enable the numerical calculation of the probabilities of each of the $2^{20}$ configurations. Figure~\ref{fig:rbm_c} shows the evolution of the $KL$ divergence between the sample distribution and the true RBM distribution in two cases: when the samples are generated from a conventional RBM using Gibbs sampling and when they are generated from the quasi-periodic network implementation. In the quasi-periodic network, sampling is done in a decentralized manner. Each neuron/unit generates an event/sample whenever one of its internal oscillators generates an event. The latest event/sample generated by each unit defines the current state of the network. A new sample is obtained as soon as each oscillator has generated at least one event since the last sample was recorded.

As the number of samples increases, the sampling distribution approaches the true RBM distribution in both cases as shown in Fig.~\ref{fig:rbm_d}. The quasi-periodic network sampling distribution eventually becomes quite close to the true RBM distribution, yet not as close as the Gibbs sampler distribution. We believe this is because the neuron/unit activation function is not exactly a logistic sigmoid (Fig.~\ref{fig:rbm_a}) which renders the quasi-periodic network distribution slightly different from the ideal distribution in Eq.~\ref{eq:rbm}. This shows, however, that a quasi-periodic event-based system can closely approximate a Gibbs sampler. 

\section{Hardware Demonstration}
\begin{figure}[t]
\label{fig:hwnw}
  \centering
  \begin{subfigure}[b]{0.38\textwidth}
    \includegraphics[width=\textwidth]{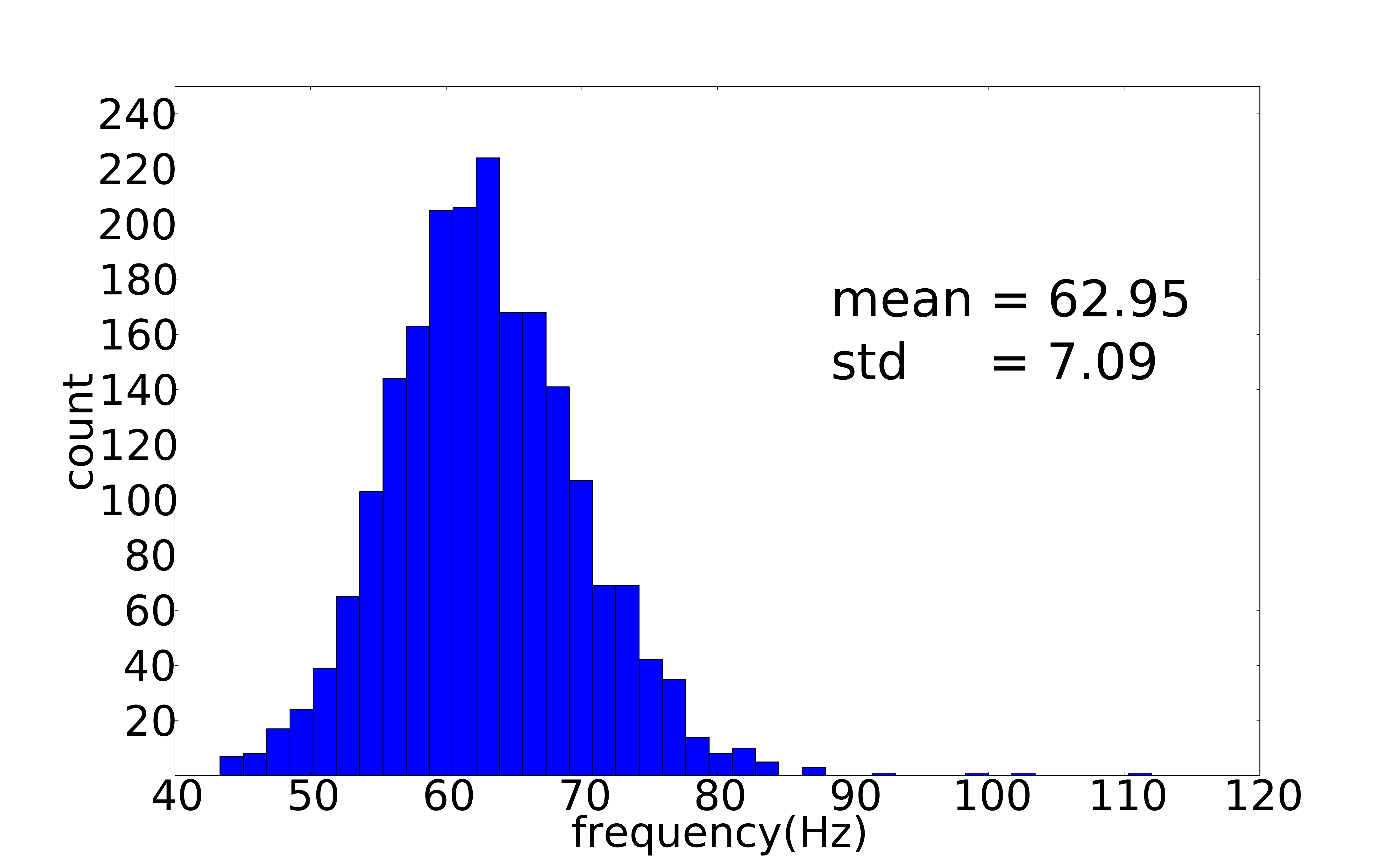}
    \subcaption{}
    \label{fig:hwnw_a}
  \end{subfigure}
  \begin{subfigure}[b]{0.2\textwidth}
    \includegraphics[width=\textwidth]{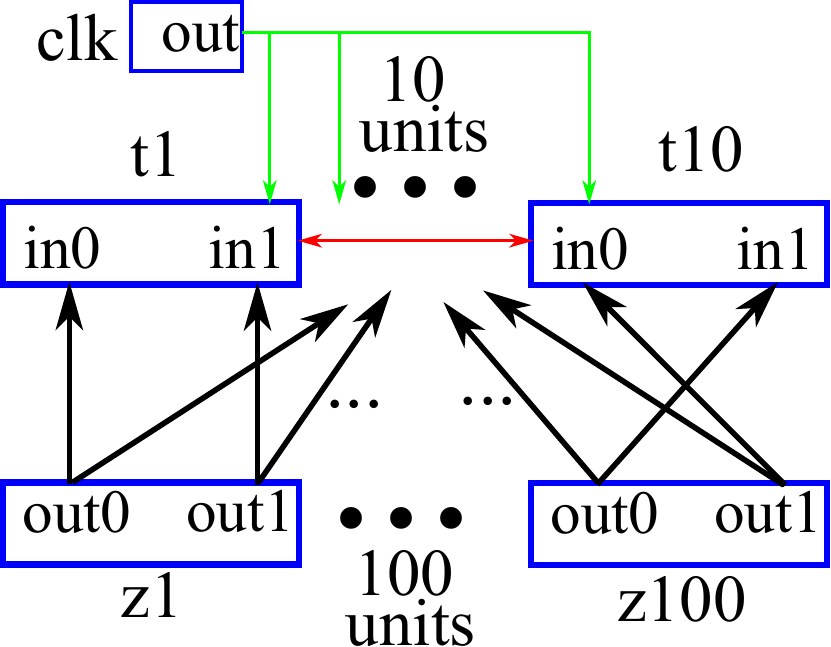}
    \subcaption{}
    \label{fig:hwnw_b}
  \end{subfigure}
  \begin{subfigure}[b]{0.38\textwidth}
    \includegraphics[width=\textwidth]{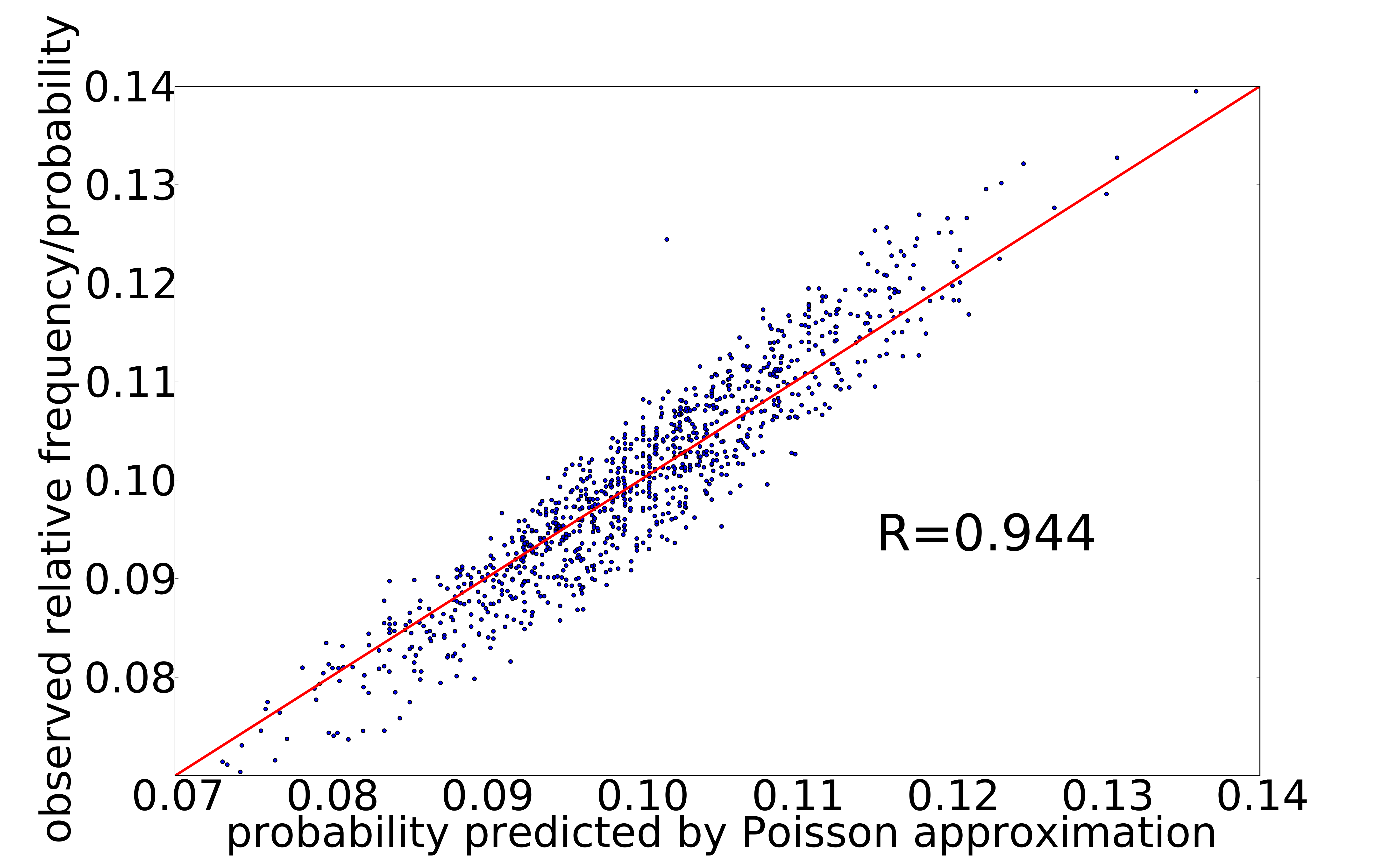}
    \subcaption{}
    \label{fig:hwnw_c}
  \end{subfigure}
\caption{(\subref{fig:hwnw_a}) Frequency distribution of the 2048 oscillators on the chip.
(\subref{fig:hwnw_b}) Structure of the network implemented on the VLSI chip. A $1$ event from a pattern unit ($t1$ to $t10$) shuts down all pattern units. A normal binary unit is designated as a clock unit ('clk') and its events reactivate the pattern units and puts them at state $0$. (\subref{fig:hwnw_c}) The observed relative frequencies of the $1$ events from each pattern unit in the chip for each input pattern compared to the predictions of Eq.~\ref{eq:hwideal}. } 
\end{figure}

We fabricated a custom chip that contains 2048 binary units/neurons in a standard $180$ nm VLSI process. Each unit has two input ports and two output ports and a 2-state FSM similar to the one shown in Fig.~\ref{fig:sigmoid_b}. The neuron binary state encodes the input port on which the last input event arrived. When the internal oscillator in a neuron generates an event, the event is routed to the output port corresponding to the current state of the neuron (see Fig.~\ref{fig:sigmoid_b}). As shown in Fig.~\ref{fig:hwnw_a}, due to transistor mismatch, the oscillator frequencies in the different units are significantly different and, since physical uncoupled analog oscillators are used, incommensurable. Fig.~\ref{fig:hwnw_b} shows the structure of the network implemented on the chip. The network has $10$ pattern units, $t1$ to $t10$. Each receives input events from $100$ input units. The uni-directional connection from an input unit to a pattern unit can either have weight $1$ (example $z1$ to $t1$) or weight $0$ (example $z100$ to $t10$). Pattern unit $i$ is thus associated with a binary weight vector ${\bf w}_i$ that defines its preferred pattern, i.e, the input unit values that will maximize its 'probability' of generating a $1$ event. 

Whenever a pattern unit generates a $1$ event, it shuts down all pattern units (including itself). An event from the 'clk' unit activates the pattern units and sets them at state $0$. Let the binary vector ${\bf z}$ denote the state of the input units. Define $m_i$ as the number of matching entries in the vectors ${\bf w}_i$ and ${\bf z}$, divided by the vector lengths ($100$). Assuming the oscillator frequencies in the different units are not very different, the number of events arriving at the $in1$ port of pattern unit $i$ in one oscillation cycle divided by the total number of received event, $x_i/N$, is on average equal to $m_i$. Assuming event generation in each unit is Poissonian instead of periodic, the probability of a pattern unit generating a one event $P(t_i=1)$ is proportional to $x_i/N = m_i$ (see the linear activation function in Fig.~\ref{fig:sigmoid_f}). But since pattern units are in a competitive configuration, only one pattern unit can generate a $1$ event for each 'clk' event, the renormalized $P(t_i=1)$ is:
\begin{equation}
\label{eq:hwideal}
P(t_i=1) = m_i/\sum\limits_{j=1}^{10}m_j
\end{equation}

In the chip experiment, each weight vector ${\bf w}_i$ was randomly initialized. $100$ different input layer patterns ${\bf z_1}-{\bf z_{100}}$ were then applied to the input layer. The events of the pattern units were collected and $P(t_i)$ evaluated from the events/samples for each input pattern. The resulting $1000$ data points are plotted in Fig.~\ref{fig:hwnw_c} as a function of the $P(t_i)$ values predicted by the Poisson assumption (Eq.~\ref{eq:hwideal}). The discrepancy is largely because the Poisson approximation assumes all units generate a Poisson train with the same rate, while the frequencies of the physical oscillators are different (Fig.~\ref{fig:hwnw_a}), thus biasing the competition in favor of pattern units with higher frequencies. If the FSM in the pattern units were a depth $3$ counter (as in Fig.~\ref{fig:sigmoid_c}), then the competition would be between a number of units whose activation functions closely approximate the logistic sigmoid and $P(t_i=1)$ would then approximately be the softmax function: $e^{m_i}/\sum\limits_{j=1}^{10}e^{m_j}$.


\section{Summary and Conclusions}
Many stochastic algorithms used in machine learning and optimization applications or as models of biological computation are formulated as a distributed stochastic network where each element integrates incoming messages/spikes, applies a stochastic non-linear transformation, then emits a message/spike. We have shown that these stochastic networks can be reformulated in a radically different way as a quasi-periodic event-based system. The combined effect of several periodic, but incommensurable, event/message streams on a target neuron with limited memory can be formulated in stochastic terms by assuming the event streams are Poissonian instead of periodic. This allows the FSM in a neuron to be treated as a Markov chain that is then used to accurately approximate the relative frequencies of the occupancies of the different FSM states in the quasi-periodic system

The scheme we describe for realizing approximations of stochastic units is quite suitable for large distributed systems as noise-generating resources in each unit are not required. By simply changing the communication scheme so that messages/events are communicated in a decentralized quasi-periodic manner, good approximation of stochastic behavior can be obtained (see the comparison to Gibbs sampling in Fig.~\ref{fig:rbm_d}). One advantage of the proposed scheme is the ease by which different approximations of stochastic activation functions can be ``programmed'', simply by changing the form of the neuron's FSM (see Fig.~\ref{fig:sigmoid_f}). We showed that the fabrication mismatch inherent in a VLSI process gives rise to incommensurable frequencies in identical oscillator circuits and the resulting quasi-periodic physical system can be used in sampling applications. By reformulating quasi-periodic event-based dynamics in stochastic terms, our results highlight a new direction for physical implementations of distributed stochastic systems.

\FloatBarrier

\bibliographystyle{unsrt}

\end{document}